\documentclass[11pt]{article}

\usepackage[]{acl}

\usepackage{times}
\usepackage{latexsym}
\usepackage[table]{xcolor}
\usepackage[T1]{fontenc}
\usepackage[utf8]{inputenc}

\usepackage{microtype}

\usepackage{inconsolata}

\usepackage{graphicx}
\usepackage{amsmath}
\usepackage{amssymb}

\usepackage{array}
\usepackage{booktabs}
\usepackage{arydshln}
\usepackage{multirow}

\title{Think Dense, Not Long: Dynamic Decoupled Conditional Advantage\\ for Efficient Reasoning}

\author{%
  Keqin Peng$^{1}$,
  Yuanxin Ouyang$^{1}$,
  Xuebo Liu$^{2}$, \\
  \textbf{Zhiliang Tian}$^{3}$,
  \textbf{Ruijian Han}$^{4}$,
  \textbf{Yancheng Yuan}$^{4}$,
  \textbf{Liang Ding}$^{5}$\thanks{~~Corresponding Authors.}\\
  $^{1}$Beihang University $^{2}$Harbin Institute of Technology, Shenzhen\\ $^{3}$National University of Defense Technology $^{4}$Hong Kong Polytechnic University $^{5}$Alibaba Group\\
\texttt{keqin.peng@buaa.edu.cn}\quad\texttt{liangding.liam@gmail.com}}

\begin{document}
\maketitle
\begin{abstract}
  Reinforcement Learning with Verifiable Rewards (RLVR) can elicit strong multi-step reasoning, yet it often encourages overly verbose traces. Moreover, naive length penalties in group-relative optimization can severely hurt accuracy. We attribute this failure to two structural issues: (i) \textbf{Dilution of Length Baseline}, where incorrect responses (with zero length reward) depress the group baseline and over-penalize correct solutions; and (ii) \textbf{Difficulty-Penalty Mismatch}, where a static penalty cannot adapt to problem difficulty, suppressing necessary reasoning on hard instances while leaving redundancy on easy ones.
  We propose \textbf{Dynamic Decoupled Conditional Advantage (DDCA)} to decouple efficiency optimization from correctness. DDCA computes length advantages \textit{conditionally} within the correct-response cluster to eliminate baseline dilution, and dynamically scales the penalty strength using the group pass rate as a proxy for difficulty. Experiments on GSM8K, MATH500, AMC23, and AIME25 show that DDCA consistently improves the efficiency--accuracy trade-off relative to adaptive baselines (e.g., \citealp{arora2025traininglanguagemodelsreason}), reducing generated tokens by approximately 60\% on simpler tasks (e.g., GSM8K) versus over 20\% on harder benchmarks (e.g., AIME25), thereby maintaining or improving accuracy. Code is available at \url{https://github.com/alphadl/DDCA}.
\end{abstract}

\section{Introduction}
\label{sec:intro}

The advent of large reasoning models (LRMs), exemplified by OpenAI's o1\footnote{\url{https://openai.com/o1/}} and DeepSeek-R1 \cite{guo2025deepseek}, has marked a paradigm shift in natural language processing from ``System 1'' rapid responses to ``System 2'' deliberate reasoning. By employing Chain-of-Thought (CoT) prompting/verification \citep{wei2022chain,wang2022selfconsistency} and Reinforcement Learning with Verifiable Rewards (RLVR), these models have achieved unprecedented success in complex mathematical and logical domains. However, this performance comes at a substantial cost: \textit{inference inefficiency}. Models trained via standard RLVR, such as Group Relative Policy Optimization (GRPO), tend to develop a behavior we term \textbf{``overthinking''}—generating excessively verbose reasoning traces that inflate computational costs and latency without necessarily improving the final answer's correctness.

Mitigating overthinking is critical for the scalable deployment of LRMs. A natural intuition is to incorporate a length penalty into the reward function to discourage verbosity. While straightforward, empirical evidence suggests that naively penalizing length often substantially degrades reasoning performance \cite{su2025thinking}. The community faces a dilemma: we can have models that are accurate but verbose, or efficient but intellectually ``shallow''.

In this work, we investigate the root cause of this trade-off. We argue that the failure of naive length penalties is not merely a hyperparameter tuning issue, but stems from structural issues in how correctness and efficiency signals interact under group-relative optimization. Through a theoretical decomposition of the GRPO objective, we identify two critical failures:
\begin{itemize}
  \item \textbf{The Dilution of Length Baseline:} In standard group-relative estimation, the baseline is computed over all samples. Since incorrect responses typically receive zero reward (and thus carry no length signal), mixed-quality groups exhibit a systematically depressed baseline. As a result, even correct solutions can be over-penalized against an artificially low threshold. \textit{DDCA addresses this by computing length advantages only within the correct-response cluster.}
  \item \textbf{Difficulty-Penalty Mismatch:} A static length penalty cannot adapt to problem difficulty. It tends to suppress necessary reasoning on hard problems (where longer traces are required for correctness), while being too weak to curb redundancy on easy problems. \textit{DDCA addresses this by scaling the efficiency pressure using the group pass rate, focusing on correctness for hard instances and efficiency for easy ones.}
\end{itemize}

To resolve these issues, we propose \textbf{Dynamic Decoupled Conditional Advantage (DDCA)}, a reward shaping framework compatible with GRPO and RLOO. DDCA disentangles efficiency from correctness with two coupled designs that map directly to the two failures above: (i) it computes length advantages \textbf{conditionally} within the correct-response cluster to eliminate baseline dilution; and (ii) it dynamically scales the penalty strength using the group pass rate to adapt length control to difficulty. We further apply a Sigmoid-based mapping to bound length signals, turning efficiency from a punitive constraint into a stable \textbf{bonus} for high information density.

We validate DDCA on the DeepSeek-R1-Distill (representing the SFT stage) and DeepScaleR (representing the further RL-refined stage) across four competitive benchmarks (GSM8K, MATH500, AMC23, AIME25). 
Experimental results demonstrate that DDCA selectively optimizes costs based on difficulty: it reduces token consumption by approx. 60\% on simpler tasks (e.g., GSM8K) while retaining 75–80\% of the reasoning length on harder benchmarks (e.g., AIME25) to ensure correctness. 
Our work shows that we can teach models to think efficiently without forcing them to think less.

Our \textbf{contributions} are summarized as follows:
\begin{itemize}
  \item We dissect the failure of length-penalized RLVR, identifying the \textit{Dilution of Length Baseline} and \textit{Difficulty-Penalty Mismatch} as the structural causes of the efficiency--accuracy trade-off.
  \item We propose DDCA, a theoretically grounded framework that decouples correctness and efficiency via conditional (correct-only) length advantages and difficulty-aware scaling.
  \item Experiments show that DDCA achieves a strong efficiency--accuracy trade-off across model families and benchmarks.
\end{itemize}

\section{Preliminaries}
\subsection{Reinforcement learning with verifiable rewards (RLVR)}
\label{subsec:RLVR}
Reinforcement learning with verifiable rewards (RLVR), notably Group Relative Policy Optimization~(GRPO; \citet{shao2024deepseekmath}), is an effective method for enhancing the reasoning ability of large language models. GRPO formulates a policy-gradient objective that removes the need for an explicit value function by using comparative advantage estimation within a group of sampled responses. Specifically, GRPO samples a group of responses $\{o_1, o_2, \ldots, o_{\text{G}}\}$ for each question, assigns each response a verifiable reward $r(\cdot)$, and then estimates its advantage using the group mean and standard deviation of rewards:
\begin{equation}
  \hat{A}_i = \frac{r({o_i}) - \text{mean}(\{r(o_1), r(o_2), \cdots, r({o_\text{G}}\}))}{\text{std}(\{r(o_1), r(o_2), \cdots, r(o_{\text{G}}\}))}.
  \label{eq:adv}
\end{equation}
Similar to PPO~\cite{schulman2017proximal}, GRPO uses a clipped objective with KL penalty and optimizes the policy model $\pi_\theta$ by maximizing the following objective:
\begin{equation}
  \begin{aligned}
    \mathcal{J}_{\text{GRPO}}(\theta) =\;&
    \mathbb{E}_{{q} \sim \mathcal{D}, \{{o}_i\}_{i=1}^{\text{G}} \sim \pi_{\theta_{\text{old}}}(\cdot|q)} \\
    &\frac{1}{\text{G}}\sum_{i=1}^{\text{G}}\frac{1}{|o_i|}\sum_{j=1}^{|o_i|}\Bigl\{
      \min\Bigl[\tau_{i,t}(\theta)\hat{A}_{i},\, \\
      &\text{clip}\bigl(\tau_{i,t}(\theta), 1-\varepsilon, 1+\varepsilon\bigr)\hat{A}_{i}\Bigr] \\
      &- \beta\,\mathbb{D}_{\text{KL}}\!\left[\pi_\theta \,\|\, \pi_{\text{ref}}\right]
    \Bigr\},
  \end{aligned}
\end{equation}
where
\begin{equation}
  \tau_{i,t} = \frac{\pi_\theta(o_{i,t}|q, o_{i,<t})}{\pi_{\theta_{\text{old}}}(o_{i,t}|q, o_{i,<t})}
\end{equation}
\begin{equation}
  \quad \mathbb{D}_{\text{KL}} (\pi_\theta \| \pi_{\text{ref}}) = \frac{1}{|o_i|} \sum_{t=1}^{|o_i|} \log \frac{\pi_\theta(o_{i,t}|q, o_{i,<t})}{\pi_{\text{ref}}(o_{i,t}|q, o_{i,<t})}.
\end{equation}
Here, $|o_i|$ is the sequence length of response $o_i$. $\pi_{\text{ref}}$ represents the reference model and the term $\mathbb{D}_{\text{KL}} (\pi_\theta \| \pi_{\text{ref}})$ indicates a KL penalty term to limit how much the trained model $\pi_\theta$ can deviate from the reference model $\pi_{\text{ref}}$.

\subsection{Rule-based Reward Modeling with Length Penalty}
For mitigating the overthinking phenomenon, one of the most straightforward methods is to add a length-based reward along with the fundamental correctness reward $R_c$ to encourage shorter yet correct responses~\citep{su2025thinking,arora2025traininglanguagemodelsreason}.
Specifically, if adopting a monotonically decreasing length reward function $R(l)=-\gamma l$ that takes the sequence length $l$ as input, the combined reward is defined as:
\begin{equation}
  R(\hat{y}, y) =
  \begin{cases}
    1 - \gamma|o_i|, & \texttt{is\_equivalent}(\hat{y}, y) \\
    0, & \text{otherwise}
  \end{cases}
  \label{eq:lp}
\end{equation}
where $y$ is the ground-truth answer and $\hat{y}$ is the predicted answer, and $\gamma$ is a small factor to prevent the length reward from leading the overall reward, which could be adaptively computed~\citep{su2025thinking} or preset as a hyperparameter~\citep{team2025kimi}.

\section{Revisiting Length Penalty in RLVR}
\label{sec:revisiting}
While introducing a length penalty is a prevalent heuristic to mitigate the overthinking phenomenon in reasoning models, empirical observations indicate that naively mixing a length penalty with correctness rewards often leads to performance degradation. In this section, we analyze the standard coupled reward formulation and identify two critical issues: the \textit{Dilution of Length Baseline} and \textit{Difficulty-Penalty Mismatch}, which collectively hinder the model's ability to balance reasoning depth and conciseness.

\subsection{The Dilution of Length Baseline}
\label{subsec: dilution}
Standard GRPO estimates the baseline using the mean reward of all samples in a group. Let a prompt $q$ generate a group of $G$ responses $\mathcal{O} = \{o_1, \dots, o_G\}$. We denote the subset of correct responses as $\mathcal{C} \subset \mathcal{O}$ with size $n$, and incorrect responses as $\mathcal{W}$ with size $G-n$. Using the reward function from Equation~\ref{eq:lp}, the group mean $\mu$ is calculated as:
\begin{equation}
  \begin{aligned}
    \mu
    &= \frac{1}{G} \sum_{i=1}^G R(y, o_i) \\
    &= \frac{1}{G} \left( \sum_{o_j \in \mathcal{C}} (1 - \gamma |o_j|) + \sum_{o_k \in \mathcal{W}} 0 \right).
  \end{aligned}
\end{equation}
Let $\bar{L}_{\mathcal{C}} = \frac{1}{n} \sum_{o_j \in \mathcal{C}} |o_j|$ be the average length of the \textit{correct} responses. The group mean simplifies to:
\begin{equation}
  \mu = \frac{n}{G} - \gamma \frac{n}{G} \bar{L}_{\mathcal{C}}.
\end{equation}
Now, consider the unnormalized advantage $A^{raw}_i = R(y, o_i) - \mu$ for a specific correct response $o_i \in \mathcal{C}$. Substituting the terms:
\begin{equation}
  \begin{aligned}
    A^{raw}_i &= (1 - \gamma |o_i|) - \left( \frac{n}{G} - \gamma \frac{n}{G} \bar{L}_{\mathcal{C}} \right) \\
    &= \left(1 - \frac{n}{G}\right) - \gamma \left( |o_i| - \frac{n}{G} \bar{L}_{\mathcal{C}} \right).
  \end{aligned}
  \label{eq:dilution_derivation}
\end{equation}
Equation~(\ref{eq:dilution_derivation}) reveals the \textbf{Dilution of Length Baseline}. Ideally, to evaluate the efficiency of $o_i$, the term $|o_i|$ should be compared against the average length of valid solutions, $\bar{L}_{\mathcal{C}}$. However, the standard formulation compares $|o_i|$ against a diluted baseline $\frac{n}{G} \bar{L}_{\mathcal{C}}$.

Since the pass rate $\frac{n}{G} \le 1$, the subtracted baseline is systematically smaller than the actual average length. This results in an exaggerated penalty term, where correct responses are punished not for being longer than peers, but for being longer than a baseline artificially lowered by incorrect (zero-reward) responses. This variance hinders the model from distinguishing between necessary reasoning and redundancy, particularly in early training stages where $n$ is small.

\subsection{Difficulty-Penalty Mismatch}
\label{subsec:difficulty}

The second issue arises from applying a static penalty coefficient $\gamma$ across problems of varying difficulty. We define the empirical difficulty of a problem inversely proportional to the group pass rate $\rho = \frac{n}{G}$.

\paragraph{High Difficulty ($\rho \to 0$):} Hard problems typically require complex reasoning chains, implying a naturally higher token count ($|o_i|$ is large). However, as derived in Eq.~(\ref{eq:dilution_derivation}), the penalty term is $-\gamma |o_i|$. If $\gamma$ is static, this negative term can dominate the reward signal for long, correct chains. Consequently, the model is incentivized to shorten responses even when extensive reasoning is critical for accuracy, potentially collapsing the reasoning chain and degrading performance on hard tasks.

\paragraph{Low Difficulty ($\rho \to 1$):} For easy problems, the model is confident, and the primary optimization objective should shift from correctness (which is already satisfied) to efficiency. A static $\gamma$ chosen to accommodate hard problems is often too lenient for easy ones, failing to sufficiently penalize redundancy.

\section{Methodology}
\label{sec:method}

To address the \textit{Dilution of Length Baseline} and \textit{Difficulty-Penalty Mismatch} identified in Section~\ref{sec:revisiting}, we propose \textbf{Dynamic Decoupled Conditional Advantage (DDCA)}. Our method decouples the optimization of reasoning correctness from token efficiency, employing a conditional RLOO estimator with dynamic difficulty scaling.

\subsection{Decoupled Advantage Framework}
Standard RLVR approaches typically mix correctness and length penalties into a single scalar reward. We propose decoupling the total advantage $A_i$ into two independent components: the accuracy advantage $A^{\text{acc}}$ and the length penalty advantage $A^{\text{len}}$.
For a prompt $x$ and a group of $N$ sampled responses $\{y_1, \dots, y_N\}$, the total advantage is formulated as:
\begin{equation}
  A_i = A_i^{\text{acc}} - \beta \cdot A_i^{\text{len}}
  \label{eq:total_advantage}
\end{equation}
where $\beta$ is a hyperparameter controlling the global penalty strength.

We employ the \textbf{Reinforce Leave-One-Out (RLOO)} estimator for both components to reduce gradient variance. The accuracy advantage is computed as:
\begin{equation}
  A_i^{\text{acc}} = r_i^{\text{acc}} - \frac{1}{N-1} \sum_{j \neq i} r_j^{\text{acc}}
\end{equation}
where $r_j^{\text{acc}} \in \{0, 1\}$ is the binary correctness label.

\subsection{Dynamic Length Reward and Advantage}
\label{subsec:dynamic_len}
Another core innovation lies in the construction of the length advantage $A_i^{\text{len}}$. This process involves three steps: (1) conditional length normalization, (2) Sigmoid reward modeling, and (3) dynamic RLOO estimation.

\paragraph{Conditional Sigmoid Reward.}
To solve the baseline dilution problem, we first compute length statistics strictly within the subset of correct responses $\mathcal{C} = \{y_j \mid r_j^{\text{acc}}=1\}$. For a correct response $y_i \in \mathcal{C}$, we calculate the Z-score of its length:
\begin{equation}
  z_i = \frac{|y_i| - \mu_{\mathcal{C}}}{\sigma_{\mathcal{C}}}, \quad \text{where } \mu_{\mathcal{C}} = \frac{1}{|\mathcal{C}|}\sum_{y_j \in \mathcal{C}}|y_j|.
\end{equation}
Instead of using the raw Z-score, we define the length reward $r_i^{\text{len}}$ using a Sigmoid transformation:
\begin{equation}
  r_i^{\text{len}} = \text{Sigmoid}(z_i) = \frac{1}{1 + e^{-z_i}}
\end{equation}
This formulation yields a bounded reward $r_i^{\text{len}} \in (0, 1)$. Responses significantly longer than peers ($z_i \gg 0$) receive a penalty reward close to 1, while concise responses approach 0. This normalization serves two purposes: it prevents extremely long "hallucinations" or short guesses (outliers) from dominating the gradient, and it ensures the penalty magnitude is comparable across different prompt lengths.

\paragraph{Dynamic RLOO Estimation.}
Finally, we compute the length advantage $A_i^{\text{len}}$ using RLOO to minimize variance. Crucially, we incorporate the \textbf{Difficulty-Aware Coefficient} $\rho = \frac{n}{N}$ (where $n=|\mathcal{C}|$) directly into this term.

For any correct response $y_i \in \mathcal{C}$, the dynamic length advantage is defined as:
\begin{equation}
  A_i^{\text{len}} = \underbrace{\left( \frac{n}{N} \right)}_{\text{Difficulty}} \cdot \underbrace{\left( r_i^{\text{len}} - \frac{1}{n-1} \sum_{j \in \mathcal{C}, j \neq i} r_j^{\text{len}} \right)}_{\text{Conditional RLOO}}
\end{equation}
For incorrect responses ($y_i \notin \mathcal{C}$), $A_i^{\text{len}} = 0$.

Combining this with Eq.~(\ref{eq:total_advantage}), our final DDCA objective adaptively modulates the optimization focus:
\begin{itemize}
  \item \textbf{Hard Problems ($n/N \to 0$):} The coefficient $\frac{n}{N}$ suppresses the length advantage. The model focuses purely on $A^{\text{acc}}$, prioritizing the discovery of correct reasoning paths regardless of length.
  \item \textbf{Easy Problems ($n/N \to 1$):} The coefficient approaches 1. The model is penalized for having a higher $r^{\text{len}}$ than its peers, driving it towards the most efficient solution within the correct cluster.
\end{itemize}

% ==================================================================
% Section 5: Experimental Setup & Results
% ==================================================================
\section{Experiments}
\label{sec:experiments}

\subsection{Experimental Setup}
\begin{sloppypar}
  \paragraph{Training.} We choose two representative reasoning models: \textbf{DeepScaleR\mbox{\,-\,}1.5B\mbox{\,-\,}Preview} and \textbf{DeepSeek\mbox{\,-\,}R1\mbox{\,-\,}Distill\mbox{\,-\,}Qwen\mbox{\,-\,}1.5B}~\citep{guo2025deepseek} as our backbone models, which have shown strong performance on mathematical reasoning tasks. Following~\citet{hou2025thinkprune}, we view them as two representative post-training paradigms. Specifically, DeepSeek-R1-Distill is obtained by supervised fine-tuning on DeepSeek-R1 outputs, which may limit its ceiling. In contrast, DeepScaleR-1.5B-Preview further applies RL on top of DeepSeek-R1-Distill-Qwen-1.5B.
\end{sloppypar}

For the training dataset, since prior work~\cite{ye2025limo} has shown that even a small but high-quality dataset can improve LLM performance via RL, we use a compact training set consisting only of historical AIME and AMC math questions. Following~\citet{hou2025thinkprune}, we use the preprocessed data from Prime~\cite{cui2025process} and take the AIME--AMC subset for training. In total, the training dataset consists of 2470 distinct training examples.

\paragraph{Evaluation.} We evaluate the models on four test datasets: GSM8K~\cite{cobbe2021training}, which contains grade-school-level math problems; MATH500~\cite{lightman2023let}, a standard benchmark with harder problems than GSM8K; AMC23; and the American Invitational Mathematics Examination (AIME) 2025, two competition-level datasets of challenging mathematical problems. For fair comparison, we set the temperature to 0.6, top-$p$ to 0.95, and use a maximum token limit of 16384 suggested by \citep{guo2025deepseek}. We conduct 4 rollouts for GSM8K and MATH500 due to their large number of test samples, and 16 rollouts for AMC23 and AIME 2025 to compute pass@1. We also compute the Average Efficiency Score~\cite{luo2025o1} for a comprehensive assessment of efficiency and efficacy.

\subsection{Metrics}
\label{sec:computation_metrics}
We use AES and pass@1 to evaluate our method.
\paragraph{AES.}
Let $p=\mathrm{pass@1}$, $p_b=\mathrm{pass@1}_{\mathrm{base}}$, $L$ be the average token cost,
and $L_b$ be the base policy token cost. The AES score~\citep{luo2025o1} is defined as:
\begin{equation}
  \label{eq:aes}
  \begin{aligned}
    \mathrm{AES} \;=\;& \frac{L_b-L}{L_b}+
    \begin{cases}
      3\,\dfrac{p-p_b}{p_b}, & p\ge p_b,\\[2pt]
      -5\,\dfrac{p_b-p}{p_b}, & p<p_b.
    \end{cases}
  \end{aligned}
\end{equation}

This metric rewards token reduction while heavily penalizing performance drops and
favoring improvements over the baseline.

\paragraph{Pass@K.} The pass@K~\citep{chen2021evaluating} scores are computed as below:
\begin{equation}
  \mathrm{pass@K}=1 - \frac{ \binom{n - c}{K} }{ \binom{n}{K} }
\end{equation}
where $n$ is the number of samples and $c$ is the number of correct samples.
When $K$ is set to 1, this metric is reduced to the average accuracy among the $n$ samples.

\paragraph{Baselines.} We compare our proposed DDCA against:
\begin{itemize}
    \setlength\itemsep{0em}
  \item \textbf{Vanilla RLOO}: The standard algorithm without explicit length constraints.
  \item \textbf{GRPO+LP}: A Standard GRPO with a linear length penalty.
  \item \textbf{ThinkPrune-4K}~\cite{hou2025thinkprune}: A length-clip reward to compress CoT length. We set the maximum length equal 4K.
  \item \textbf{TLMRE}~\cite{su2025thinking}: An efficient training method with non-linear length penalty.
\end{itemize}

% ==================================================================
% Table 1: Main Results (Restored & Populated)
% ==================================================================

\begin{table*}
  \centering
  \scalebox{0.85}{
    \begin{tabular}{cccccccccccc}
      \toprule
      \textbf{Model} & \multicolumn{2}{c}{\textbf{GSM8K}} & \multicolumn{2}{c}{\textbf{MATH500}} & \multicolumn{2}{c}{\textbf{AMC23}} & \multicolumn{2}{c}{\textbf{AIME25}} & \multicolumn{3}{c}{\textbf{Average}}            \\
      \cmidrule(lr){2-3}\cmidrule(lr){4-4}\cmidrule(lr){5-5}\cmidrule(r){6-7}\cmidrule(lr){8-9}\cmidrule(lr){10-12}
      & \textbf{Acc.} & \textbf{\#Tok.}    & \textbf{Acc.} & \textbf{\#Tok.}      & \textbf{Acc.} & \textbf{\#Tok.}    & \textbf{Acc.} & \textbf{\#Tok.}     & \textbf{Acc.} & \textbf{\#Tok.} & \textbf{AES}  \\
      \hline\hline
      \multicolumn{12}{c}{\textit{DeepSeek-R1-Distill-1.5B}}                                                                                                                                                                                   \\
      \midrule
      \textbf{Base}                   & 81.2          & 1683                & 85.0          & 4662                 & 71.8          & 7945               & 21.9          & 12158               & 65.0          & 6612            & 0.00          \\
      \textbf{+RLOO}                  &  85.1         & 1921               & 86.1          & 4166                 & 76.1          & 6953               & 25.0          & 11202                & 68.1          & 6061            & 0.23          \\
      \hdashline
      \textbf{+LP}                    & 76.1            &  421                 &  84.3           & 3069                    &  74.5           &  5336                 &  23.3           &  10160                  &  64.6           &  4747              &  0.25            \\
      \textbf{ThinkPrune-4k}         & 83.2            & 1138                  & 85.5            & 3235                    & 72.8            & 5107                  & 25.2            & 8879                   & 66.7            & 4590               & 0.38             \\
      \textbf{TLMRE}                  & 84.3          & 1826                & 85.5          & 4209                 & 72.2          & 7319               & 25.0          & 11465                & 66.8          & 6205            &  0.14         \\
      \textbf{Ours}              & 85.0          & 1211                & 86.2          & 3248                 & 76.3          & 5556               & 26.5          & 9616                & 68.5          & 4908            &   0.42        \\
      \hline\hline
      \multicolumn{12}{c}{\textit{DeepScaleR-1.5B-Preview}}                                                                                                                                                                                                 \\
      \midrule
      \textbf{Base}                   & 86.9          & 1700                & 90.1          & 3093                 & 81.7          & 4993              & 31.1          & 8301               & 72.5         &  4522  & 0.00          \\
      \textbf{+RLOO}                  & 87.2          & 1424               & 89.5          & 2843                 & 82.3          & 4468               & 30.8          & 7994                & 72.5          & 4182            &  0.08         \\
      \hdashline
      \textbf{+LP}                    & 77.2            & 224                  & 89.1            & 2036                   & 79.4            & 3226                  & 27.5            & 6437                   & 68.3            & 2981               & 0.05             \\
      \textbf{ThinkPrune-4k}         & 87.1            & 1433                  & 89.2            & 2641                    & 81.3            & 4179                  & 30.8            & 7044                   & 72.1            & 3824               & 0.13             \\
      \textbf{TLMRE}                  & 86.4          & 785                & 89.3          & 2324                 & 79.1          & 3734               & 29.2          & 6879                &  71.0         &   3431          &   0.14        \\
      \textbf{Ours}              & 86.9          & 672                & 88.8          & 2082                 & 83.2          & 3515               & 29.8          & 6109                & 72.2          & 3095            &  0.30         \\
      \bottomrule
    \end{tabular}
  }
  \caption{\textbf{Main Results.} Comparison of accuracy (\%), average token consumption and Accuracy-Efficiency Score (AES). DDCA achieves better AES across different datasets and backbones.}
  \label{tab:main_results}
\end{table*}

\section{Main Results}
\label{src:results}
We evaluate \textit{DDCA} on four reasoning benchmarks spanning elementary to competition-level difficulty (GSM8K, MATH500, AMC23, AIME25) and two backbone models (DeepSeek-R1-Distill-1.5B and DeepScaleR-1.5B-Preview). Table~\ref{tab:main_results} summarizes the accuracy--efficiency trade-off against representative baselines. 

\textbf{DDCA improves the efficiency--accuracy trade-off.}
As shown in Table~1, \textbf{Ours (DDCA)} consistently achieves the highest Accuracy-Efficiency Score (AES) across both model backbones (0.42 for DeepSeek-R1 and 0.30 for DeepScaleR). Compared to the strong \textit{+RLOO} baseline on DeepSeek-R1-Distill-1.5B, DDCA not only improves average accuracy by \textbf{0.4\%} ($68.1\% \rightarrow 68.5\%$) but also substantially reduces average token consumption by approximately \textbf{20\%} ($6061 \rightarrow 4908$). Even on the already optimized DeepScaleR-1.5B-Preview, DDCA reduces token usage by over \textbf{1,000 tokens} on average ($4182 \rightarrow 3095$) while maintaining comparable accuracy ($72.2\%$ vs. $72.5\%$). This confirms that decoupling the length advantage and applying conditional normalization allows the model to cut redundancy without sacrificing reasoning capabilities.

\paragraph{DDCA preserves reasoning capacity on hard tasks.}
The benefits of our dynamic difficulty-aware mechanism are most pronounced on challenging benchmarks. Standard length penalties (\textit{+LP}) often degrade performance on hard tasks (e.g., dropping AIME25 accuracy from 25.0\% to 23.3\% on DeepSeek-R1) by prematurely truncating necessary reasoning chains. In contrast, DDCA achieves the \textbf{highest accuracy of 26.5\%} on AIME25 with the DeepSeek backbone, outperforming both \textit{TLMRE} (25.0\%) and \textit{ThinkPrune-4k} (25.2\%). This validates that scaling the penalty by the pass rate ($\frac{n}{N}$) effectively relaxes constraints for difficult problems, preserving the "deep thinking" capacity required for competition-level mathematics while aggressively optimizing efficiency on easier tasks like GSM8K (where tokens are reduced to $\sim$1200).

\section{Analysis}
We further conduct a comprehensive analysis to investigate the properties of DDCA from three perspectives: \textbf{capability}, \textbf{mechanism}, and \textbf{sensitivity}. 
First, we evaluate the method's capabilities by examining its scalability across varying token budgets (Section~\ref{subsec:scalability}) and its ability to maintain exploration diversity (Section~\ref{subsec:exploration}). 
Next, we validate the underlying mechanisms, quantifying the specific mitigation of baseline dilution (Section~\ref{subsec:dilution_analysis}) and verifying the contribution of each design component through ablation studies (Section~\ref{subsec:ablation}). 
Finally, we analyze the hyperparameter sensitivity regarding the penalty coefficient $\beta$ to provide practical usage insights (Section~\ref{subsec:beta_sensitivity}).

\subsection{Scalability across Token Budgets}
\label{subsec:scalability}

\begin{figure}[t]
  \centering
  \includegraphics[width=\linewidth]{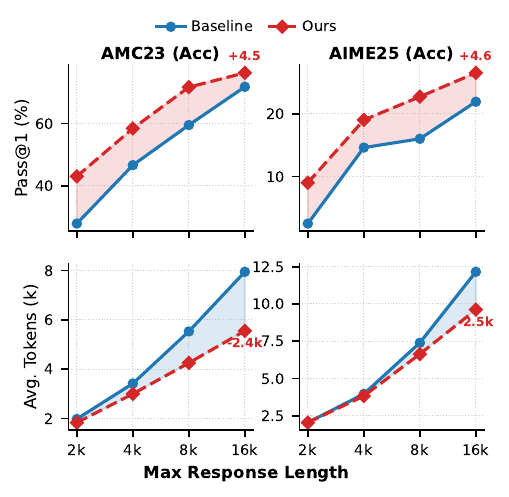}
  \caption{\textbf{Performance and efficiency scaling.}
  \textbf{Top:} Pass@1 accuracy. DDCA (red) outperforms Baseline (blue) on both AMC23 and AIME25) tasks.
\textbf{Bottom:} Token consumption. DDCA exhibits a plateauing trend, saving up to 2.5k tokens on AIME25 at the 16k budget.}
\label{fig:single_col_results}
\end{figure}

We evaluate DDCA across varying token budgets (2k to 16k). Figure~\ref{fig:single_col_results} presents the results on AMC23 and AIME25.

As shown in Figure~\ref{fig:single_col_results}, DDCA consistently improves accuracy while curbing token usage.
On the challenging AIME25 dataset (right column), the dynamic difficulty coefficient ($\frac{n}{N}$) allows the model to utilize deeper reasoning when needed, resulting in a $+4.6\%$ accuracy gain at 16k budget, while still reducing the average computational cost by maintaining conciseness on solved sub-paths.

\subsection{Exploration Capability Analysis}
\label{subsec:exploration}

\begin{figure}[t]
\centering
% \linewidth 在单栏中约为 3.3-3.5 英寸
\includegraphics[width=\linewidth]{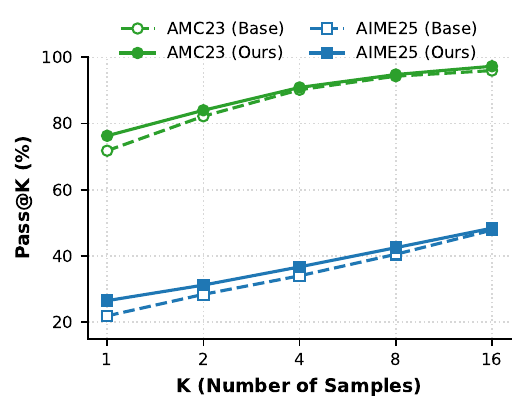}
\caption{\textbf{Pass@K comparison.} Solid lines represent DDCA (Ours), and dashed lines represent the Baseline. DDCA consistently outperforms the baseline across all $K$ values on both AMC23 (Green) and AIME25 (Blue), demonstrating that our dynamic penalty efficiently prunes redundancy without compromising the diversity of valid reasoning paths.}
\label{fig:pass_k}
\end{figure}

We further investigate whether the length penalty in DDCA hinders the model's exploration capability. Figure~\ref{fig:pass_k} presents the Pass@K performance ($K \in \{1, \dots, 16\}$).

As shown in Figure~\ref{fig:pass_k}, DDCA (solid lines) maintains a consistent performance margin over the Baseline (dashed lines). Crucially, on the hard AIME25 benchmark, the slope of our Pass@K curve remains steep. This indicates that DDCA does not suffer from mode collapse; instead, the dynamic coefficient $\frac{n}{N}$ successfully acts as a regularizer, encouraging the model to explore diverse yet concise reasoning trajectories.

\subsection{Mitigation of Baseline Dilution}
\label{subsec:dilution_analysis}

To empirically verify that DDCA mitigates the \textit{Dilution of Length Baseline}, we analyzed the performance on the MATH-500 benchmark, stratified by problem difficulty (Levels 1-5).
Baseline dilution occurs when incorrect responses (with effectively zero length penalty contribution) pull down the group mean, distorting the advantage estimation. This phenomenon is theoretically most severe in hard problems where the pass rate is low (i.e., mixed-quality groups are common).

\begin{figure}[t]
\centering
\includegraphics[width=\linewidth]{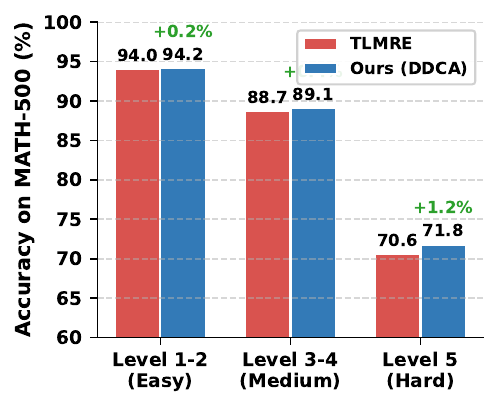}
\caption{\textbf{Accuracy breakdown on MATH-500 by difficulty.} While performance gains are marginal on easy problems (Level 1-2) where baseline dilution is negligible, DDCA achieves a significant \textbf{+1.2\% boost} on hard problems (Level 5). This confirms that decoupling the advantage is crucial for robust optimization in low-pass-rate regimes.}
\label{fig:math500_breakdown}
\end{figure}

The results in Figure~\ref{fig:math500_breakdown} confirm our hypothesis:
\begin{itemize}
\item \textbf{Level 1-2 (Easy):} The pass rates are high ($\sim94\%$), meaning groups consist almost entirely of correct responses. Here, baseline dilution is negligible, and DDCA performs comparably to TLMRE ($94.2\%$ vs. $94.0\%$).
\item \textbf{Level 5 (Hard):} The pass rates drop to $\sim70\%$, introducing significant variance in response quality within groups. Standard methods like TLMRE suffer here because the length baseline is diluted by incorrect answers. In contrast, DDCA decouples the efficiency estimation, comparing correct solutions only against other correct ones. Consequently, we observe the largest performance gain of \textbf{+1.2\%} ($70.6\% \to 71.8\%$) in this high-difficulty regime.
\end{itemize}
This demonstrates that our decoupled advantage formulation effectively robustifies the reward signal against the noise introduced by incorrect samples, a critical property for solving complex reasoning tasks.

\subsection{Component Ablation Study}
\label{subsec:ablation}
To verify the distinct contributions of the \textit{Decoupled Advantage} and the \textit{Dynamic Difficulty Coefficient} ($\frac{n}{N}$), we examine the performance of DDCA variants on both MATH (easy) and AIME25 (hard) datasets. The results are summarized in Table~\ref{tab:ablation}.

\paragraph{Impact of Dynamic Coefficient (w/o Dynamic).}
The "w/o Dynamic" variant applies a static decoupled penalty. While it achieves the lowest token usage on MATH-500 dataset, it suffers significantly on the hard AIME25 task, yielding a low accuracy of 21.9\%. This confirms the \textit{Difficulty-Penalty Mismatch}: a static penalty indiscriminately truncates the extensive reasoning chains required for complex problems. In contrast, the full DDCA utilizes the dynamic term to relax the penalty for hard tasks, restoring accuracy to \textbf{26.5\%}.

\paragraph{Impact of Decoupling (w/o Decouple).}
The "w/o Decouple" variant incorporates the difficulty coefficient but calculates advantage across the mixed group (including incorrect responses). On the easy MATH dataset, it fails to significantly reduce token consumption compared to the baseline ($4662 \to 3973$). This validates the existence of \textit{Baseline Dilution}: the inclusion of incorrect responses lowers the reward baseline, weakening the efficiency signal. DDCA resolves this by decoupling the baseline, successfully compressing the reasoning path.

\paragraph{Conclusion.}
DDCA effectively synergizes both components. The \textit{Decoupling} mechanism ensures precise efficiency optimization for solvable problems, while the \textit{Dynamic} mechanism acts as a safety guard to preserve reasoning depth for difficult problems.

\begin{table}[t]
\centering
\resizebox{\linewidth}{!}{
  \begin{tabular}{lcccc}
    \toprule
    \multirow{2.5}{*}{\textbf{Method}} & \multicolumn{2}{c}{\textbf{Math (Easy)}} & \multicolumn{2}{c}{\textbf{AIME25 (Hard)}} \\
    \cmidrule(lr){2-3} \cmidrule(lr){4-5}
    & \textbf{Acc} ($\uparrow$) & \textbf{Tokens} ($\downarrow$) & \textbf{Acc} ($\uparrow$) & \textbf{Tokens} ($\downarrow$) \\
    \midrule
    Baseline & 85.0 & 4662 & 21.9 & 12158 \\
    DDCA & 86.2 & 3248 & \textbf{26.5} & 9616 \\
    \hdashline
    -w/o Dynamic (Static) & 86.3 & \textbf{3044} & 21.9 & \textbf{8992} \\

    -w/o Decouple (Coupled) & \textbf{86.5} & 3973 & 25.4 & 11111 \\
    \bottomrule
  \end{tabular}
}
\vspace{2pt}
\caption{\textbf{Ablation study of component effectiveness.} Comparison of DDCA variants on MATH and AIME25. \textit{w/o Dynamic} suffers from low accuracy on hard tasks due to excessive penalization. \textit{w/o Decouple} fails to optimize efficiency on easy tasks due to baseline dilution. \textbf{DDCA} achieves the best trade-off.}
\label{tab:ablation}
\end{table}

\subsection{Sensitivity Analysis of Penalty Coefficient}
\label{subsec:beta_sensitivity}

\begin{figure}[t]
\centering
\includegraphics[width=\linewidth]{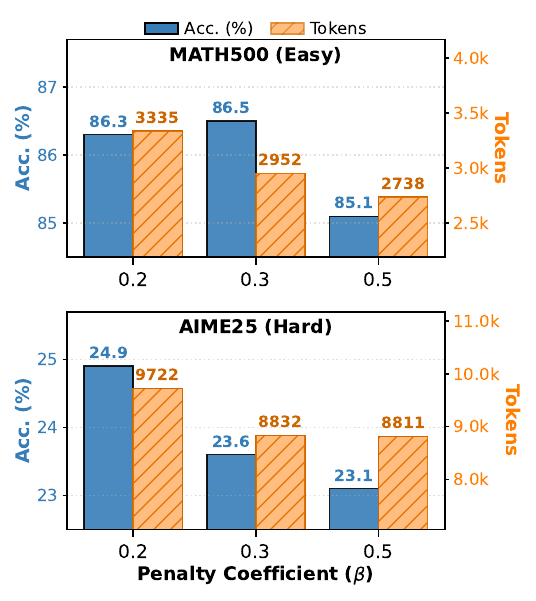}
\caption{\textbf{Effect of $\beta$ on Accuracy and Token Usage.}
  \textbf{Top:} On MATH500, a moderate $\beta=0.3$ boosts accuracy to $86.5\%$ while reducing tokens.
  \textbf{Bottom:} On AIME25, accuracy drops as $\beta$ increases ($24.9\% \to 23.1\%$), showing high sensitivity.
Blue bars: Accuracy; Orange hatched bars: Token count.}
\label{fig:beta_optimized}
\end{figure}

We investigate the trade-off between reasoning accuracy and token efficiency by varying the penalty coefficient $\beta \in \{0.2, 0.3, 0.5\}$. Figure~\ref{fig:beta_optimized} illustrates the results on MATH500 and AIME25.

\paragraph{Trade-off Analysis.}
The results demonstrate a clear dichotomy. On the easier MATH500 benchmark, increasing $\beta$ from 0.2 to 0.3 improves accuracy by removing redundant reasoning steps ("overthinking"). However, on the harder AIME25 benchmark, any increase in $\beta$ strictly degrades accuracy, as the penalty truncates the extensive exploration required for complex solutions. In all cases, token usage (Orange) decreases monotonically with $\beta$. This justifies our dynamic mechanism: applying stricter penalties on easy tasks while relaxing them on hard ones.

\section{Related Work}
\begin{sloppypar}
\paragraph{Reinforcement Learning For Reasoning.} Reinforcement learning (RL) is widely used for post-training large language models, especially to improve multi-step reasoning. Alignment-oriented post-training methods such as RLHF have become standard recipes \citep{ouyang2022training,rafailov2023dpo,garg2025ipo}, and reasoning-oriented post-training has been revitalized by long-thinking LRMs such as OpenAI’s o1\footnote{\url{https://openai.com/o1/}} and DeepSeek-R1 \citep{guo2025deepseek}. Reinforcement Learning with Verifiable Rewards (RLVR) further enables learning from automatically checkable outcomes (e.g., math and code) \citep{code-r1}, with algorithmic advances such as GRPO \citep{shao2024deepseekmath} and DAPO \citep{yu2025dapo}. Complementary analyses also show that deeper problem understanding can boost math reasoning performance \citep{zhong2024dup}, and explicit reasoning chains can improve robustness in related settings \citep{wu-etal-2025-reasonke}. Beyond RL-based post-training, in-context learning performance is also sensitive to demonstration selection strategies \citep{peng-etal-2024-revisiting}. Our work targets a distinct issue: efficiency constraints can distort group-relative RLVR advantages; we address this via advantage-level reward shaping to preserve correctness while controlling length.

\paragraph{Length Penalty for Efficient Reasoning.}
Overly long reasoning traces (\textit{overthinking}) increase inference costs \citep{chen2024not} and may be driven by repeated re-verification (\textit{self-doubt}) \citep{peng2025selfdoubt}. A range of approaches have been explored \citep{zhu2025towards}, with a common direction integrating length awareness into RL. For example, \citet{arora2025traininglanguagemodelsreason} ranks sampled solutions by correctness and normalized length, while Song et al.~\cite{concisewalk} improve GRPO (GRPO++/L-GRPO) with additional mechanisms for stable training. Unlike these coupled or static formulations, DDCA decouples efficiency from correctness at the advantage level by computing length advantages only among correct responses and scaling efficiency pressure by pass rate, directly addressing baseline dilution and difficulty mismatch.
\end{sloppypar}

\section{Conclusion}
In this work, we revisit the length penalty in RLVR and identify the \textit{Dilution of Length Baseline} and \textit{Difficulty-Penalty Mismatch} as two fundamental flaws in standard group-relative optimization. We show that coupled rewards inevitably lead to either uncontrollable length collapse in easy tasks or the suppression of reasoning in hard tasks. We propose \textbf{Dynamic Decoupled Conditional Advantage (DDCA)}, which decouples the efficiency objective and applies it conditionally to correct responses. Extensive experiments demonstrate that DDCA achieves a superior Pareto frontier, reducing token consumption by up to \textbf{60\%} on easier benchmarks while preserving the "deep thinking" capacity required for hard tasks (e.g., AIME25). 
Our findings highlight that efficient reasoning requires incentivizing higher information density within valid reasoning paths, rather than simply penalizing length. We hope DDCA serves as a robust baseline for future research in efficient System 2 scaling.

\section*{Limitations}
While our proposed DDCA demonstrates significant improvements in reasoning efficiency, we acknowledge several limitations:
\begin{itemize}
\item \textbf{Domain Specificity:} Our theoretical analysis and experiments focus primarily on mathematical and logical reasoning tasks (STEM), where "correctness" is binary and verifiable. The applicability of DDCA to open-ended generation or creative writing tasks, where the definition of "efficiency" is more subjective, remains to be explored.
\item \textbf{Hyperparameter Sensitivity:} Although DDCA is more robust than vanilla length penalties, the efficiency coefficient $\beta$ still requires tuning. An overly aggressive $\beta$ might still encourage the model to skip necessary verification steps in extremely hard problems, albeit to a lesser extent than baseline methods.
\end{itemize}

\bibliography{custom}

@string{CoRR = "CoRR"}

@string{arXiv = "arXiv"}

@string{NeurIPS = "Adv. Neural Inf. Process. Syst. (NeurIPS)"}

@string{ICLR = "Int. Conf. Learn. Represent. (ICLR)"}

@article{schulman2017proximal,
  title={Proximal policy optimization algorithms},
  author={Schulman, John and Wolski, Filip and Dhariwal, Prafulla and Radford, Alec and Klimov, Oleg},
  journal={arXiv preprint arXiv:1707.06347},
  year={2017},
  doi={10.48550/arXiv.1707.06347},
  url={https://arxiv.org/abs/1707.06347}
}

@article{guo2025deepseek,
  title={Deepseek-r1: Incentivizing reasoning capability in llms via reinforcement learning},
  author={Guo, Daya and Yang, Dejian and Zhang, Haowei and Song, Junxiao and Zhang, Ruoyu and Xu, Runxin and Zhu, Qihao and Ma, Shirong and Wang, Peiyi and Bi, Xiao and others},
  journal=arXiv,
  volume={abs/2501.12948},
  year={2025},
  doi={10.48550/arXiv.2501.12948},
  url={https://arxiv.org/abs/2501.12948}
}

@article{su2025thinking,
  title={Thinking Fast and Right: Balancing Accuracy and Reasoning Length with Adaptive Rewards},
  author={Su, Jinyan and Cardie, Claire},
  journal=arXiv,
  volume={abs/2505.18298},
  year={2025},
  doi={10.48550/arXiv.2505.18298},
  url={https://arxiv.org/abs/2505.18298}
}

@inproceedings{arora2025traininglanguagemodelsreason,
  title={Training Language Models to Reason Efficiently},
  author={Arora, Daman and Zanette, Andrea},
  booktitle=NeurIPS,
  year={2025},
  note={arXiv:2502.04463},
  doi={10.48550/arXiv.2502.04463},
  url={https://neurips.cc/virtual/2025/poster/119443}
}

@article{team2025kimi,
  title={Kimi k1.5: Scaling Reinforcement Learning with LLMs},
  author={Team, Kimi and Du, Angang and Gao, Bofei and Xing, Bowei and Jiang, Changjiu and Chen, Cheng and Li, Cheng and Xiao, Chenjun and Du, Chenzhuang and Liao, Chonghua and others},
  journal=arXiv,
  volume={abs/2501.12599},
  year={2025},
  doi={10.48550/arXiv.2501.12599},
  url={https://arxiv.org/abs/2501.12599}
}

@article{cobbe2021training,
  title={Training verifiers to solve math word problems},
  author={Cobbe, Karl and Kosaraju, Vineet and Bavarian, Mohammad and others},
  journal={arXiv preprint arXiv:2110.14168},
  year={2021},
  doi={10.48550/arXiv.2110.14168},
  url={https://arxiv.org/abs/2110.14168}
}

@inproceedings{lightman2023let,
  title={Let's verify step by step},
  author={Lightman, Hunter and Kosaraju, Vineet and Burda, Yuri and Edwards, Harrison and Baker, Bowen and Lee, Teddy and Leike, Jan and Schulman, John and Sutskever, Ilya and Cobbe, Karl},
  booktitle={International Conference on Learning Representations (ICLR)},
  year={2023},
  url={https://openreview.net/forum?id=O1gRv0T4uX}
}

@article{luo2025o1,
  title={O1-pruner: Length-harmonizing fine-tuning for o1-like reasoning pruning},
  author={Luo, Haotian and Shen, Li and He, Haiying and Wang, Yibo and Liu, Shiwei and Li, Wei and Tan, Naiqiang and Cao, Xiaochun and Tao, Dacheng},
  journal={arXiv preprint arXiv:2501.12570},
  year={2025},
  doi={10.48550/arXiv.2501.12570},
  url={https://arxiv.org/abs/2501.12570}
}

@misc{code-r1,
  title={Code-R1: Reproducing R1 for Code with Reliable Rewards},
  author={Liu, Jiawei and Zhang, Lingming},
  howpublished={\url{https://github.com/ganler/code-r1}},
  year={2025},
  note={GitHub repository},
  url={https://github.com/ganler/code-r1}
}

@article{shao2024deepseekmath,
  title={Deepseekmath: Pushing the limits of mathematical reasoning in open language models},
  author={Shao, Zhihong and Wang, Peiyi and Zhu, Qihao and Xu, Runxin and Song, Junxiao and Bi, Xiao and Zhang, Haowei and Zhang, Mingchuan and Li, YK and others},
  journal={arXiv preprint arXiv:2402.03300},
  year={2024},
  doi={10.48550/arXiv.2402.03300},
  url={https://arxiv.org/abs/2402.03300}
}

@article{yu2025dapo,
  title={DAPO: An Open-Source LLM Reinforcement Learning System at Scale},
  author={Yu, Qiying and Zhang, Zheng and Zhu, Ruofei and Yuan, Yufeng and Zuo, Xiaochen and Yue, Yu and Dai, Weinan and Fan, Tiantian and Liu, Gaohong and Liu, Lingjun and others},
  journal={arXiv preprint arXiv:2503.14476},
  year={2025},
  doi={10.48550/arXiv.2503.14476},
  url={https://arxiv.org/abs/2503.14476}
}

@article{chen2024not,
  title={Do not think that much for 2+ 3=? on the overthinking of o1-like llms},
  author={Chen, Xingyu and Xu, Jiahao and Liang, Tian and He, Zhiwei and Pang, Jianhui and Yu, Dian and Song, Linfeng and Liu, Qiuzhi and Zhou, Mengfei and Zhang, Zhuosheng and others},
  journal={arXiv preprint arXiv:2412.21187},
  year={2024},
  doi={10.48550/arXiv.2412.21187},
  url={https://arxiv.org/abs/2412.21187}
}

@article{zhu2025towards,
  title={Towards concise and adaptive thinking in large reasoning models: A survey},
  author={Zhu, Jason and Li, Hongyu},
  journal={arXiv preprint arXiv:2507.09662},
  year={2025},
  doi={10.48550/arXiv.2507.09662},
  url={https://arxiv.org/abs/2507.09662}
}

@article{concisewalk,
  title={Walk Before You Run! Concise LLM Reasoning via Reinforcement Learning},
  author={Song, Mingyang and Zheng, Mao},
  journal={arXiv preprint arXiv:2505.21178},
  year={2025},
  doi={10.48550/arXiv.2505.21178},
  url={https://arxiv.org/abs/2505.21178}
}

@article{chen2021evaluating,
  title={Evaluating large language models trained on code},
  author={Chen, Mark},
  journal={arXiv preprint arXiv:2107.03374},
  year={2021},
  doi={10.48550/arXiv.2107.03374},
  url={https://arxiv.org/abs/2107.03374}
}

@inproceedings{wei2022chain,
  title={Chain-of-Thought Prompting Elicits Reasoning in Large Language Models},
  author={Wei, Jason and Wang, Xuezhi and Schuurmans, Dale and Bosma, Maarten and Ichter, Brian and Xia, Fei and Chi, Ed and Le, Quoc V. and Zhou, Denny},
  booktitle={Advances in Neural Information Processing Systems (NeurIPS)},
  year={2022},
  volume={35},
  doi={10.48550/arXiv.2201.11903},
  url={https://papers.neurips.cc/paper_files/paper/2022/hash/9d5609613524ecf4f15af0f7b31abca4-Abstract-Conference.html},
  note={arXiv:2201.11903}
}

@inproceedings{wang2022selfconsistency,
  title={Self-Consistency Improves Chain of Thought Reasoning in Language Models},
  author={Wang, Xuezhi and Wei, Jason and Schuurmans, Dale and Le, Quoc V. and Chi, Ed H. and Narang, Sharan and Chowdhery, Aakanksha and Zhou, Denny},
  booktitle={International Conference on Learning Representations (ICLR)},
  year={2023},
  doi={10.48550/arXiv.2203.11171},
  url={https://iclr.cc/virtual/2023/poster/11718},
  note={arXiv:2203.11171}
}

@article{ouyang2022training,
  title={Training language models to follow instructions with human feedback},
  author={Ouyang, Long and Wu, Jeff and Jiang, Xu and Almeida, Diogo and others},
  journal={Computing Research Repository (CoRR)},
  volume={abs/2203.02155},
  year={2022},
  doi={10.48550/arXiv.2203.02155},
  url={https://arxiv.org/abs/2203.02155},
  note={arXiv:2203.02155}
}

@article{rafailov2023dpo,
  title={Direct Preference Optimization: Your Language Model is Secretly a Reward Model},
  author={Rafailov, Rafael and Sharma, Archit and Mitchell, Eric and Ermon, Stefano and Manning, Christopher D. and Finn, Chelsea},
  journal={Computing Research Repository (CoRR)},
  volume={abs/2305.18290},
  year={2023},
  doi={10.48550/arXiv.2305.18290},
  url={https://arxiv.org/abs/2305.18290},
  note={arXiv:2305.18290}
}

@article{garg2025ipo,
  title={IPO: Your Language Model is Secretly a Preference Classifier},
  author={Garg, Shivank and Singh, Ayush and Singh, Shweta and Chopra, Paras},
  journal={Computing Research Repository (CoRR)},
  volume={abs/2502.16182},
  year={2025},
  doi={10.48550/arXiv.2502.16182},
  url={https://arxiv.org/abs/2502.16182},
  note={arXiv:2502.16182}
}

@article{hou2025thinkprune,
  title={ThinkPrune: Pruning Long Chain-of-Thought of LLMs via Reinforcement Learning},
  author={Hou, Bairu and Zhang, Yang and Ji, Jiabao and Liu, Yujian and Qian, Kaizhi and Andreas, Jacob and Chang, Shiyu},
  journal={Computing Research Repository (CoRR)},
  volume={abs/2504.01296},
  year={2025},
  doi={10.48550/arXiv.2504.01296},
  url={https://arxiv.org/abs/2504.01296},
  note={arXiv:2504.01296}
}

@article{peng2025selfdoubt,
  title={Revisiting Overthinking in Long Chain-of-Thought from the Perspective of Self-Doubt},
  author={Peng, Keqin and Ding, Liang and Ouyang, Yuanxin and Fang, Meng and Tao, Dacheng},
  journal={arXiv preprint arXiv:2505.23480},
  year={2025},
  month={May},
  doi={10.48550/arXiv.2505.23480},
  url={https://arxiv.org/abs/2505.23480}
}

@inproceedings{wu-etal-2025-reasonke,
  title={Robust Knowledge Editing via Explicit Reasoning Chains for Distractor-Resilient Multi-Hop {QA}},
  author={Wu, Yuchen and Ding, Liang and Shen, Li and Tao, Dacheng},
  booktitle={Findings of the Association for Computational Linguistics: {EMNLP} 2025},
  month={nov},
  year={2025},
  pages={14578--14586},
  publisher={Association for Computational Linguistics},
  url={https://aclanthology.org/2025.findings-emnlp.786/}
}

@article{ye2025limo,
  title={Limo: Less is more for reasoning},
  author={Ye, Yixin and Huang, Zhen and Xiao, Yang and Chern, Ethan and Xia, Shijie and Liu, Pengfei},
  journal={arXiv preprint arXiv:2502.03387},
  year={2025},
  url={https://arxiv.org/pdf/2502.03387}
}

@article{cui2025process,
  title={Process reinforcement through implicit rewards},
  author={Cui, Ganqu and Yuan, Lifan and Wang, Zefan and Wang, Hanbin and Zhang, Yuchen and Chen, Jiacheng and Li, Wendi and He, Bingxiang and Fan, Yuchen and Yu, Tianyu and others},
  journal={arXiv preprint arXiv:2502.01456},
  year={2025},
  doi={10.48550/arXiv.2502.01456},
  url={https://arxiv.org/abs/2502.01456}
}

@article{zhong2024dup,
  title={Achieving >97\% on {GSM8K}: Deeply Understanding the Problems Makes {LLMs} Better Reasoners},
  author={Zhong, Qihuang and Wang, Kang and Xu, Ziyang and Liu, Juhua and Ding, Liang and Du, Bo and Tao, Dacheng},
  journal={Computing Research Repository (CoRR)},
  volume={abs/2404.14963},
  year={2024},
  doi={10.48550/arXiv.2404.14963},
  url={https://arxiv.org/abs/2404.14963}
}

@inproceedings{peng-etal-2024-revisiting,
    title = "Revisiting Demonstration Selection Strategies in In-Context Learning",
    author = "Peng, Keqin  and
      Ding, Liang  and
      Yuan, Yancheng  and
      Liu, Xuebo  and
      Zhang, Min  and
      Ouyang, Yuanxin  and
      Tao, Dacheng",
    booktitle = "Proceedings of the 62nd Annual Meeting of the Association for Computational Linguistics",
    month = aug,
    year = "2024",
    address = "Bangkok, Thailand",
    publisher = "Association for Computational Linguistics",
    url = "https://aclanthology.org/2024.acl-long.492/",
    doi = "10.18653/v1/2024.acl-long.492",
    pages = "9090--9101",
}

@inproceedings{sheng2025hybridflow,
  title={Hybridflow: A flexible and efficient rlhf framework},
  author={Sheng, Guangming and Zhang, Chi and Ye, Zilingfeng and Wu, Xibin and Zhang, Wang and Zhang, Ru and Peng, Yanghua and Lin, Haibin and Wu, Chuan},
  booktitle={Proceedings of the Twentieth European Conference on Computer Systems},
  pages={1279--1297},
  year={2025},
  url={https://arxiv.org/pdf/2409.19256}
}

@misc{slime_github,
  author       = {Zilin Zhu and Chengxing Xie and Xin Lv and slime Contributors},
  title        = {slime: An LLM post-training framework for RL Scaling},
  year         = {2025},
  howpublished = {\url{https://github.com/THUDM/slime}},
  note         = {GitHub repository. Corresponding author: Xin Lv},
  urldate      = {2025-06-19}
}

\appendix
\section{Implementation Details}
We build on the Verl codebase~\cite{sheng2025hybridflow}\footnote{We also implement our method upon slime~\cite{slime_github} in our code.}. For all experiments, we utilize four RTX-5090 GPUs on a single low-density node. We set vLLM to maximum context length (16K) during generation and set the generation temperature to 0.6. For training, we set the maximum response equal 8192 and generate 8 responses for each prompt. For every iteration, 8 prompts are selected from the dataset and the global batch size is
set to 32. For the 1.5B, the learning rate is set to 2e-6, and the value of the KL coefficient is set to 1e-3. We use the same prompt template for all models, which can be found below:

\vspace{0.5em} 

\noindent 
\texttt{\$QUESTION. Please reason step by step, and put your final answer within \textbackslash boxed\{\}.} \\

\end{document}